\documentclass[11pt,a4paper]{article}
\usepackage[hyperref]{acl2021}
\usepackage{times}
\usepackage{latexsym}

\usepackage{microtype}
\usepackage{graphicx}
\usepackage{subfigure}
\usepackage{amsfonts,amssymb}
\usepackage{threeparttable}
\usepackage{amsfonts}
\usepackage{nicefrac}
\usepackage{booktabs}
\usepackage{url}

\title{Math-KG: Construction and Applications of  Mathematical \\ Knowledge Graph}

\author{Jianing Wang\thanks{ \quad Corresponding Author.} \\
  East China Normal University\\
  \texttt{lygwjn@gmail.com}
}

\date{}

\begin{document}
\maketitle
\begin{abstract}
Recently, the explosion of online education platforms makes a success in encouraging us to easily access online education resources. However, most of them ignore the integration of massive unstructured information, which inevitably brings the problem of \textit{information overload} and \textit{knowledge trek}. In this paper, we proposed a mathematical knowledge graph named Math-KG, which automatically constructed by the pipeline method with the natural language processing technology to integrate the resources of the mathematics. It is built from the corpora of Baidu Baike, Wikipedia. We implement a simple application system to validate the proposed Math-KG can make contributions on a series of scenes, including faults analysis and semantic search. The system is publicly available at GitHub \footnote{\url{https://github.com/wjn1996/Mathematical-Knowledge-Entity-Recognition}.}.
\end{abstract}


\section{Introduction}

Online education (OE) is a novel teaching and learning concept, which makes full use of Internet technology to realize education intelligence. By applying artificial intelligence (AI) technology in the teaching process, it can provide students with personalized learning guidelines and improve teaching efficiency. In the context of the rapid development of AI technology, the concept of intellectual adaptation has begun to be integrated into the field of education.

However, the number of online education platforms is becoming bigger and bigger, most of them ignore the integration of unstructured information, which limit to a simple classification and display of resources. In this case, it inevitably brings the problem of \textit{information overload} and \textit{knowledge trek}. For example, in mathematics, massive knowledge points provided by different platforms are independent of each other, which results in the inconsistency of knowledge and causes the confusion of students' learning process. In addition, little work concentrates on the integration of educational resources, because the field of education contains a large number of subjects, and each of them has great differences.

\begin{figure*}[ht]
\centering\includegraphics[height=52mm]{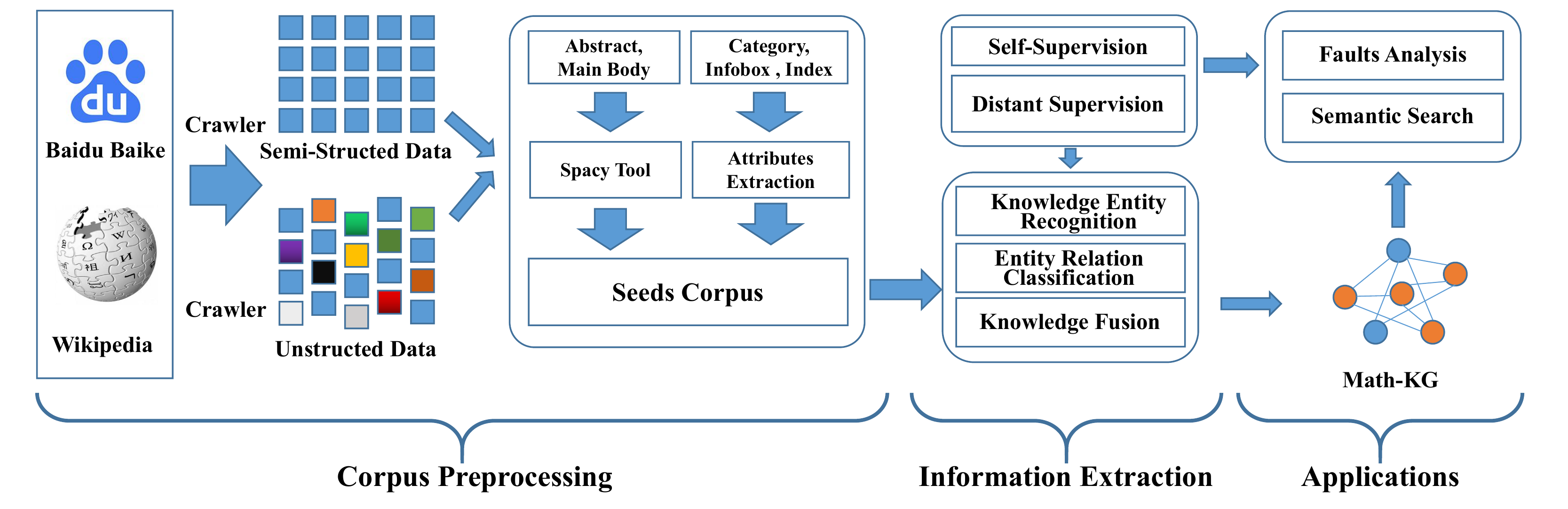}
\caption{Construction and applications of Math-KG}
\label{fig1}
\end{figure*}%

The knowledge Graph was originally proposed by Google Inc.  in 2012 to enable precise question and answer and semantic search engines by allowing computers to understand the semantics of text. The knowledge Graph aims to constructs the knowledge base by extracting a series of triples from unstructured or semi-structured texts. At present, many advanced knowledge extraction tools \cite{manning-etal-2014-stanford,gardner-etal-2018-allennlp,li-etal-2020-gaia} are proposed to construct knowledge graph, which has been widely used in the fields of medicine \cite{Goodwin2013Automatic}, financial \cite{Fu2018Stochastic}, agricultural \cite{Xiang2017Research}, etc., but it is still a blank in mathematical education.

In order to further promote the development of online mathematical education, we propose a knowledge graph-based Math-KG to integrate knowledge resources. Specifically, we define two entity types and six relation types at first, and then propose a pipeline method for Math-KG construction. We obtain the unstructured text corpus through the web crawler and created the dataset via distant supervision strategy. We utilize the natural language processing technology tool to realize the named entity recognition, entity relation classification and knowledge fusion, and then save them in the graph database. The specific construction process as described in section \ref{section2}. We implement a simple application system to validate the feasibility of knowledge graph, including faults analysis and semantic search. The main contributions of this paper are as follows:

\begin{itemize}
    \item We propose a pipeline method of mathematical knowledge graph construction, and build a large-scale knowledge base named Math-KG.
    \item We conduct some experiments to evaluate each part of our pipeline method, and implement a simple application system to make validation of some educational scenes, such as faults analysis and semantic search.
\end{itemize}

\section{Method of Math-KG Construction}
\label{section2}
As illustrated in Fig \ref{fig1}, the pipeline of Math-KG construction contains two components: \textbf{Corpus Preprocessing} and \textbf{Information Extraction}.

\subsection{Corpus Preprocessing}
\label{section2.1}

In order to construct the Math-KG, it is necessary to complete the acquisition and preprocessing of the mathematical corpus. The preprocessing stages mainly include the following parts:

\textbf{Resources Acquisition}. We crawl semi-structured and unstructured data \footnote{For the pictures and videos, we only extract the hyperlinks as an attribution} from web resources includes Baidu Baike\footnote{\url{https://baike.baidu.com/}.}, Wikipedia\footnote{\url{https://www.wikidata.org}.}, etc. Specifically, for one entity token, we capture five resources consists of categories, abstracts, main bodies, infobox blocks and corresponding indexes.

\textbf{Resources Process}. For the categories, infobox blocks and corresponding indexes, we direct view them as attributes of the entity token. The other resources contain raw text and formulas. The texts are split into sentences and processed by Spacy tool\footnote{\url{https://spacy.io/}.} which consists of word tokenization, part-of-speech (POS) tagging, dependency tree parse (DTP), etc. The mathematical formulas are transformed into latex characters by MathOCR\footnote{\url{https://sourceforge.net/projects/mathocr/}.}. We release our preprocessing code on GitHub \footnote{\url{https://github.com/wjn1996/scrapy_for_zh_wiki}.}.

\begin{table*}\small
\centering
\begin{tabular}{|p{20mm}|p{50mm}|p{50mm}|p{20mm}|}
\toprule
\textbf{relation}& \textbf{instruction}& \textbf{example} & \textbf{entity pairs}\\
\midrule
dependencies& two knowledge points have sequential dependence in the learning process  & \textit{Learning \textbf{addition}$_{e_1}$ helps us understand the definition of \textbf{subtraction}$_{e_2}$.} & Dep($e_1$,$e_2$) \\
\midrule
affiliation& two knowledge points have containing and be contained relations &\textit{An \textbf{isosceles triangle}$_{e_1}$ is a special kind of \textbf{triangle}$_{e_2}$.} & Aff($e_1$,$e_2$) \\
\midrule
equivalence& two entities in term of different name have the same meaning & \textit{\textbf{orthogon}$_{e_1}$ belongs to plane geometry which called \textbf{rectangle}$_{e_2}$}. & Equ($e_1$,$e_2$) \\
\midrule
opposite& two  knowledge points are conceptually opposite & \textit{if a number is \textbf{even}$_{e_1}$, it must not be \textbf{odd}$_{e_2}$}. & Ant($e_1$,$e_2$) \\
\midrule
synonyms& two  knowledge points have similar meanings & \textit{Circles are both \textbf{axial symmetric}$_{e_1}$ and centrally \textbf{symmetric figure}$_{e_2}$}. & Syn($e_1$,$e_2$) \\
\midrule
has properties& one knowledge point is a property of another & \textit{the \textbf{area}$_{e_1}$ of a \textbf{parallelogram}$_{e_2}$ is base times height}. & Pro($e_1$,$e_2$) \\
\bottomrule 
\end{tabular}
\caption{Instruction and example of six relations}
\label{tb1}
\end{table*}

\subsection{Information Extraction}
\label{section2.2}

Information extraction aims to extract structure triple and construct a large-scale graph. It consists of knowledge entity recognition, entity relation classification and knowledge fusion.

\textbf{Knowledge Entity Recognition}. Knowledge entity recognition can be seen as a sequence labeling task. We define two types of knowledge entities, \textbf{Conceptual Entity} and \textbf{Theorem Entity}. The \textbf{Concept Entity} indicates that the knowledge point is an existence concept, which is represented by \textit{CON}, such as \textit{triangle}, \textit{trigonometric function}, etc. The \textbf{Theorem Entity} represents the theorem, method or rule based on the  \textbf{Concept Entity} , which labeled as \textit{LEG}, for example, the entity \textit{pythagorean theorem}, which can be used to solve the \textit{hypotenuse of right triangle}, belongs to \textbf{Theorem Entity}. In order to improve the accuracy of identification, we prefix the first character of each entity with 'B' and mark the rest with 'I', other non-entities are tagged as 'O'. We first use the entries in the encyclopedia as seeds and recall some high-quality entities by heuristic rules, such as tf-idf, pattern matching or word2vec \cite{Mikolov2013Distributed} similarity. Then, we construct a small dataset via distant supervision \cite{mintz2009distant} method by aligning the corpus with the recall entities set and fine-tune by the BERT model\cite{2018BERT}.

\textbf{Entity Relation Classification}. We consider relation extraction as a classification task, which is a prerequisite for building triples. We define six classes of relation for knowledge points, namely: \textit{dependencies}, \textit{affiliation}, \textit{equivalence}, \textit{antisense}, \textit{synonyms}, \textit{has properties}. (Table \ref{tb1} gives the instruction and example of each relation.) These relations have two directions so that Math-KG will be composed of 12 kinds of edge with an NA \footnote{NA denotes that it has no relation between two entities.}. To recognize the relation between entities, we first use the explicit triples as seeds, which from categories, infobox blocks and corresponding indexes that directly provided by Baidu Baike and Wikipedia. Then, we defined some pattern rules, and use the self-supervision method to realize pattern discovery to recall some triples. In order to improve the quality of the seeds, we manually performed a rough inspection of them. Although the rule-based method can obtain high accuracy, it still lacks generalization ability. Therefore, we follow distant supervision strategy \cite{mintz2009distant} to build a large-scale dataset by heuristically aligning with raw text, and then train by BERT model\cite{2018BERT} to make the classification. However, distant supervision assumes that all the text with the same entity pair can express the same relation, which brings many noisy labeling data \cite{mintz2009distant}. To alleviate this problem, we add a sentence-level attention layer on the top of the BERT encoder and use the embeddings of special [CLS] token to make the classification.

\textbf{Knowledge Fusion}. In order to improve the accuracy of the knowledge graph, we used entity linker TagMe \footnote{\url{https://sobigdata.d4science.org/group/tagme}.} and reference resolution technique Neuralcoref \footnote{\url{https://github.com/huggingface/neuralcoref}.} to align the constructed triples with some existing knowledge bases to improve confidence. In addition, we unify these resources manually to ensure the consistency of knowledge.

\begin{table}
\centering
\setlength\tabcolsep{7pt}
\begin{tabular}{c|cccc}
\hline
 Data set & train & dev & test & all \\
\hline
\textbf{KER} & 6265 & 1300 & 1385 & 8950 \\
\textbf{ERC} & 9356 & 2000 & 2009 & 13365 \\
\hline
\end{tabular}
\caption{The statistics of knowledge entity recognition (KER) and entity relation classification (ERC) data set.}
\label{statistics}
\end{table}

\begin{table}
\centering
\setlength\tabcolsep{7pt}
\begin{tabular}{c|ccc}
\hline
 Data set & precision \% & recall \% & F1 \% \\
\hline
\textbf{KER} & 92.40 & 92.85 & 92.62 \\
\textbf{ERC} & 84.30 & 85.90 & 85.09 \\
\hline
\end{tabular}
\caption{The evaluation results of knowledge entity recognition (KER) and entity relation classification (ERC) data set.}
\label{evaluation}
\end{table}

\begin{figure*}
\centering
\subfigure[Faults Analysis]{
\includegraphics[width=52mm,height=48mm]{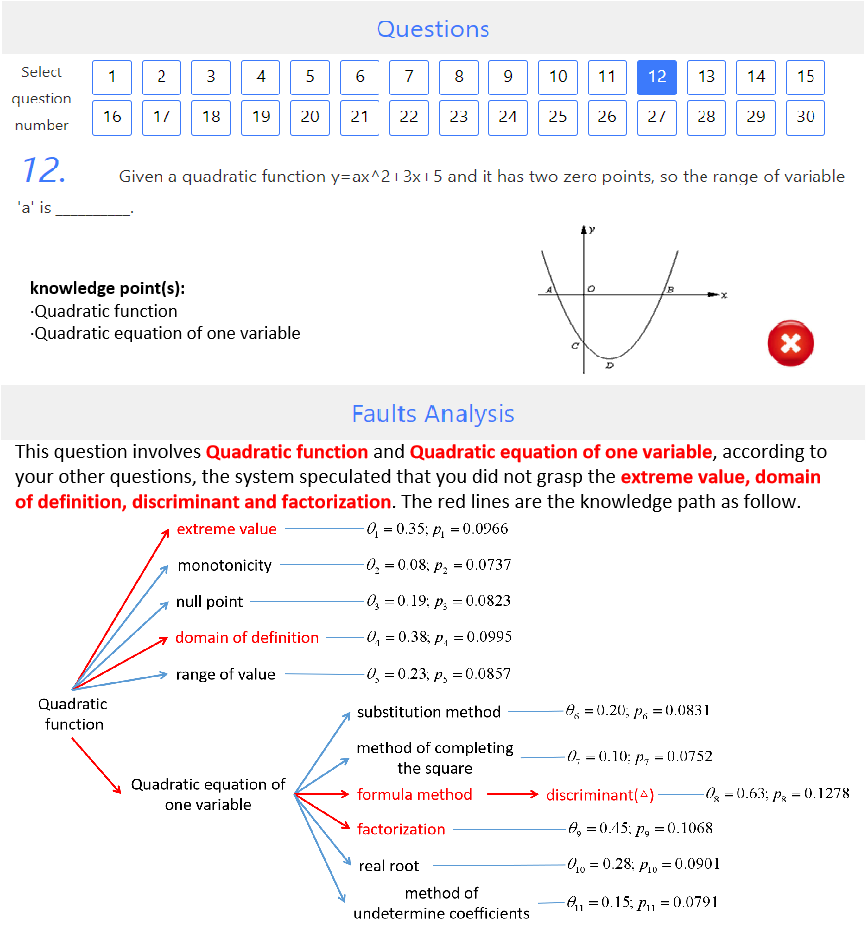}
}
\quad
\subfigure[Semantic Search]{
\includegraphics[width=93mm,height=48mm]{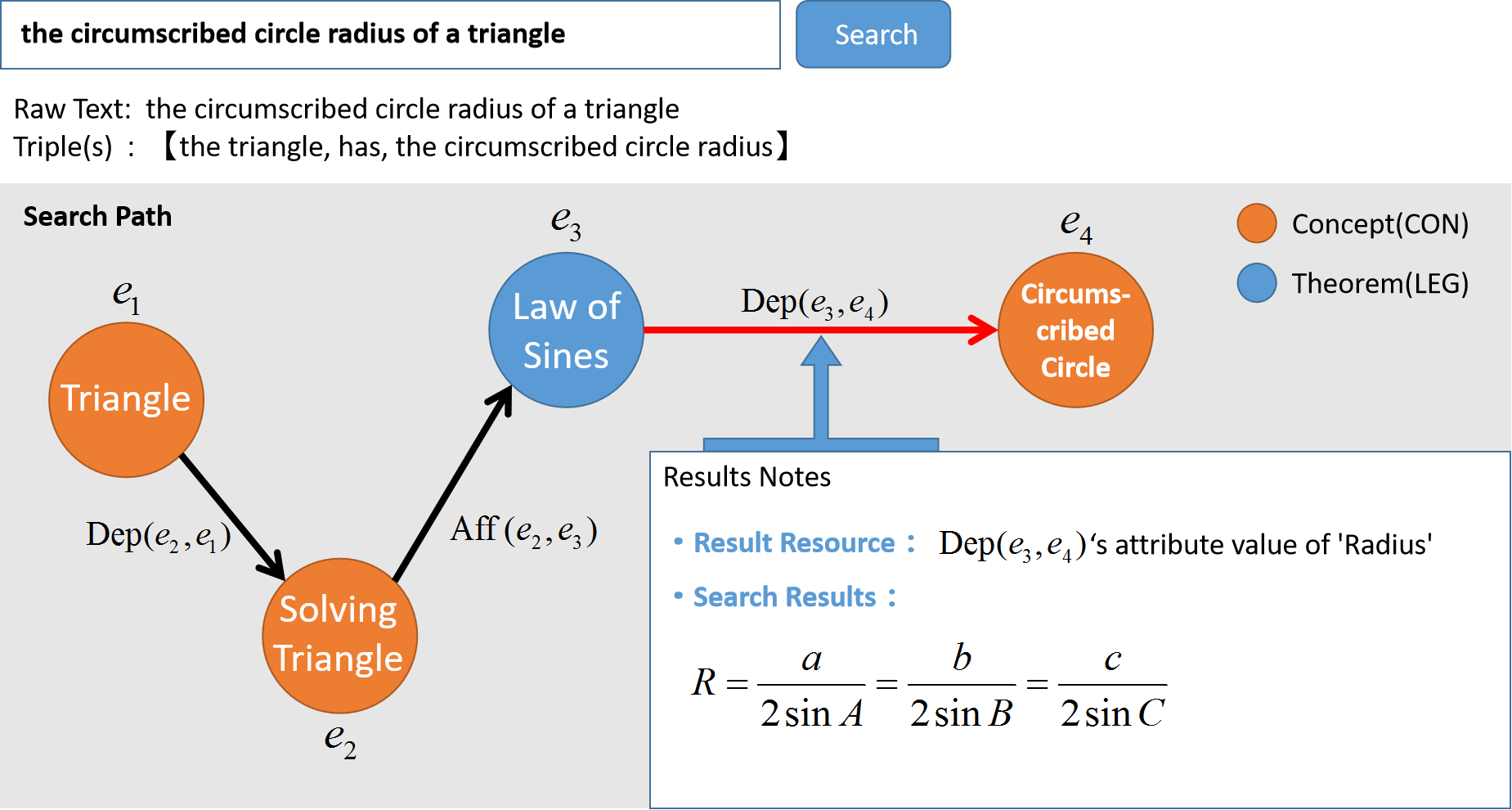}
}
\caption{Applications demonstration}
\label{fig2}
\end{figure*}

\section{Evaluations of Math-KG}
\label{section3}

\subsection{Experimental Evaluations}
\label{section3.1}

We release two core data set and corresponding Tensorflow codes \footnote{the KER and ERC data set and corresponding codes released on the github: \url{https://github.com/wjn1996/Mathematical-Knowledge-Entity-Recognition}.} to support the pipeline of construction, including knowledge entity recognition (KER) and entity relation classification (ERC). The statistics of each data set as shown in Table \ref{statistics}. We use BERT+CRF model to train on the KER data set. We use BERT+Attention+Softmax to train on the ERC data set. We use large-based BERT for both two experiments, and the batch size is set as 64, the epoch is set as 30. We choose cross-entropy loss function to train. The F1 values of KER and ERC are 92.62\% and 85.09\%, respectively. The results indicates that the current algorithm still has room for improvement. At last, we obtain 1905 mathematical knowledge point entities and 8,019 triples. Among them, there are 1337 concept entities and 568 theorem entities. The number of triples with relation \textit{dependencies}, \textit{affiliation}, \textit{equivalence}, \textit{antisense}, \textit{synonyms} and \textit{has properties} are 2016, 2609, 76, 301, 493 and 2524, respectively.

\subsection{Application Evaluations}
\label{section3.2}

In order to validate the usefulness of Math-KG, we design two application scenes, such as faults analysis and semantic search.

\textbf{Faults Analysis}. Faults analysis aims to speculate the knowledge points that student fails to master by analyzing the historical question answer results. We use the baidu, google search engines to extract some mathematics practices. We design a simple method to make faults analysis based on historical answer results. For one student, we count the number of correct and incorrect questions for each knowledge point. If the number of incorrect questions is bigger than the correct, this student fails to master this knowledge. Therefore, we can obtain a sub-graph for one practice, which contains other related knowledge points. If one knowledge point often appears in the sub-graph, it can be regarded as the source of the faults, and then be recommended to the student. As shown in Fig \ref{fig2}(a), the question labeled as two knowledge consists of \textit{Quadratic function} and \textit{Quadratic equation of one variable}. In order to show more intuition, we transform the sub-graph of these knowledge points into a tree, where the root denotes the current knowledge point of the practice, and the leaf node denotes the related knowledge points. We calculate the score based on the statistics at each leaf to represent the relevance to the root knowledge point. The few red paths represent the inference evidences.

\textbf{Semantic Search}. We also provide semantic search, which is also a research scope called knowledge base question answering (KBQA) \cite{tan-etal-2016-improved}. As we all known, knowledge graph can be viewed as a semantic network \cite{Chen2019Convolutional}, which can be answered the question by multi-hop inferences. Especially, given a question $s$, we use the Spacy tool to recognize the corresponding topic entity and recall the multi-hop sub-graph from Math-KG. We pre-trained our Math-KG by TransE \cite{Fan2014Transition} to obtain the embeddings of each knowledge entity. Then, we calculate the similarity based on a simple neural network between the question and each candidate entities, and obtain the top results. We display the path information from the topic entity to the resulting entity. As shown in Fig \ref{fig2}(b), when querying \textit{the circumscribed circle radius of a triangle}, the system recognizes the topic entity \textit{triangle} and find a multi-hop path (\textit{the triangle}, \textit{has}, \textit{the circumscribed circle radius}).

\section{Conclusion}
\label{section4}

In this paper, in order to alleivate the problem of \textit{information overload} and \textit{knowledge trek} in mathematics, we propose a pipeline method of Math-KG construction to integrate the mathematical knowledge resources. We conduct experiments to evaluate the performance of construction, and implement two applications to validate the usefulness of Math-KG, which provides students with faults analysis and semantic search. In future researches, we will put our efforts into improving the accuracy and efficiency of information extraction, and apply it to other disciplines, including physics, chemistry, etc.

\bibliographystyle{acl_natbib}
\bibliography{anthology,acl2021}


\end{document}